# General Purpose (GenP) Bioimage Ensemble Combining New Data Augmentation Techniques and Handcrafted Features

**Loris Nanni [1],* Sheryl Brahnam[2], Stefano Ghidoni[1], Gianluca Maguolo[1] and Michelangelo Paci[3]**

[1] DEI, Via Gradenigo 6, 35131 Padova, Italy; loris.nanni@unipd.it, gianluca.maguolo@unipd.it, and stefano.ghidoni@unipd.it
[2] Department of Information Technology and Cybersecurity, Missouri State University, 901 S, National Street, Springfield, MO 65804, USA; sbrahnam@missouristate.edu
[3] Tampere University, Arvo Ylpön katu 34, FI-33520, Tampere, Finland; michelangelo.paci@tuni.fi
* Correspondence: loris.nanni@unipd.it

**Abstract:** Bioimage classification plays a crucial role in many biological problems. In this work, we present a new General Purpose (GenP) ensemble that boosts performance by combining local features, dense sampling features, and deep learning approaches. First, we introduce three new methods for data augmentation based on PCA/DCT; second, we show that different data augmentation approaches can boost the performance of an ensemble of CNNs; and, finally, we propose a set of handcrafted/learned descriptors that are highly generalizable. Each handcrafted descriptor is used to train a different Support Vector Machine (SVM), and the different SVMs are combined with the ensemble of CNNs. Our method is evaluated on a diverse set of bioimage classification problems, covering tissue sections, cells, and subcellular structures, etc. Results demonstrate that the proposed GenP bioimage ensemble obtains state-of-the-art performance without any ad-hoc dataset tuning of the hyperparameters of the ensemble: the same ensemble is used for all the datasets. The CNNs are trained using the different training sets.

**Keywords:** Microscopy imaging; Data Augmentation; Deep Learning; Support Vector Machine; Convolutional Neural Networks.

## 1. Introduction

### 1.1. Background and Driving Forces

Biomedical research is increasingly dependent on computer vision and machine learning in the discovery of new knowledge and methods of diagnosis. Storing, retrieving, and analyzing large biological images has become critical, in part because of the enormous amounts of data generated by recent advances in imaging technologies. Automated image analysis has become an indispensable tool for handling the throve of data being collected for a variety of applications such as cell phenotype recognition, subcellular localization, and histopathological classification [1-4].

Because computer vision and image classification rely on methods for extracting highly discriminative feature sets, a significant area of research in these domains has focused on generating ever better methods for extracting robust descriptors. Until recently, however, most bioimage research has concentrated on the problem of segmentation [5] with little attention devoted to investigating the discriminative power of texture descriptors–even though it has been shown that extracting highly discriminative texture descriptors can circumvent the problem of segmentation [5-7]. Some whole image methods of note that have been proposed in the literature include [8-10], and some popular descriptors used in automatic bioimage classification include traditional Gabor filters [11] and Haralick's famous texture features [12].

More recent descriptors applied to automatic bioimage analysis include such powerful handcrafted descriptors as the scale-invariant feature transform (SIFT) and local binary patterns (LBP) [13-15]. Handcrafted descriptors are designed by researchers to extract specific image characteristics and is typically accomplished as follows: characteristic regions of an image are located by a key point detector, and these regions are described by a vector of measurements (the descriptor) that depends on the specific image characteristics under consideration. The extracted set of descriptors is then used to train a classifier, such as the Support Vector Machine (SVM) [16].

In contrast to handcrafted descriptors are learned descriptors, which are automatically learned by a classifier system. Learned descriptors have only recently been explored in bioimage classification. In [17], for instance, an automatic feature discovery method was proposed using class-specific dictionaries for the diagnosis of ovarian carcinomas, and [18] handcrafted and learned descriptors were combined to discriminate irregularities in brain cells; this system joined an unsupervised feature learning method with learned linear combinations of Riesz wavelets.

Within the last few years, some innovative learned approaches have been proposed that exploit deep learners [19], esp. the Convolutional Neural Network (CNN). It appears that deep learners analyze input images via the different layers in the architecture by evaluating sets of features learned directly from observations of the training images [20]. Some layers are even thought to preprocess images using a pyramidal approach [21]. When deep neural networks, such as CNN, are trained on a set of images for a specific classification problem, features extracted by the shallowest layers (close to the classification layer) strongly depend on the training set, while features taken from the first layers resemble Gabor filters or color blobs that are transferable to many other classification problems [22]. This discovery has been exploited by bioimage researchers who have used CNN or ensembles of CNNs as feature extractors in this domain [23, 24]; the resulting learned features are then treated like SIFT and LBP and become the input to traditional classifiers, such as SVM.

It is well known that when it comes to deep architecture, large image datasets are needed. The requirement of large datasets is difficult to meet in some fields: this is certainly the case in bioimaging where images are expensive to acquire and patient availability is limited. In such contexts, data augmentation is a popular method for increasing images in small datasets: indeed, such techniques have been used extensively in the analysis of medical and biological images [25]. Data augmentation also prevents overfitting CNNs and is known to improve performance. A good example of this approach in bioimage classification can be found in [26], where the researchers were able to accurately detect breast cancer in a set of histology images containing less than 100 images per class by combining pre-trained deep network architectures with multiple augmentation techniques.

The most common methods of image data augmentation involve reflection, translation, and rotation [27-30], as these augmentations generate different representations of the same sample. Different representations of a given image can also be constructed by altering the contrast, saturation, and brightness [27, 29, 30] of images. Yet another common technique is PCA jittering, which accentuates the most relevant features of an image by adding to it some of its principal components multiplied by a small number [27, 29]. Most deep learning frameworks make use of a limited set of basic image transforms. Recently, however, libraries of fast augmentation methods have been developed, such as Albumentations [31], that provide large sets of image transforms with easy-to-use wrappers around other augmentation libraries.

Specific problem-dependent augmentation methods can also be applied to expand small datasets. For example, [32], the authors replicated speckle noise, a common artifact in synthetic aperture radar images, by applying random pointwise multiplications to images, and [33], different stretchings of the human body were reproduced by creating elastic deformations of breast cancer images. Operations like elastic transforms and grid distortions are useful in medical imaging, where non-rigid structures that have shape variations are quite common [31].

In [34] an ensemble of networks for histopathological biopsy was recently proposed where the authors successfully employed multiple data augmentation techniques to improve their accuracy. Their augmentation protocol included horizontal and vertical flips, contrast enhancement, zoom, shear, and rotations. In [35] color data augmentation for the segmentation of histological images was also proposed that combined the color structure of an image with the appearance of another one.

Another method for enlarging small datasets uses Generative Adversarial Networks (GANs) to synthesize new images that are different from those contained in the original dataset [36-39]. GANs are based on an adversarial game between two neural networks: a generator network that produces synthetic samples given a perturbation and a discriminator network that distinguishes between the generator's synthesized image and the true image. Because GANs generate new images on a separately trained network, they produce, unlike data augmentation techniques, a unique yet relevant set of new images. In [40] a new GAN augmentation was proposed the output of which was smooth thanks to a spatial constraint. The authors in [40] also managed to create synthetic annotations for microscopy segmentation.

### *1.2. Research Goals and Contributions*

In this work, we present a new General Purpose (GenP) bioimage classification method that combines both handcrafted and learned descriptors. Ideally, GenP image classification systems can handle a broad range of different image classification tasks within a specific domain (such as medical imagery) with little (if any) parameter tuning. A GenP system should also perform competitively against other systems that are optimized for task-specific image classification problems.

As illustrated in Figure 1, our latest bioimage GenP system combines handcrafted features and deep learning methods to obtain a high degree of generalizability across a range of bioimage datasets. As in [41], the system proposed here takes a representative set of robust handcrafted descriptors and individually trains them on separate SVMs, with the set of SVMs combined by sum rule. Similarly, we also generate a high-performing ensemble of CNNs that is then combined with the SVM ensemble for a final decision.

Unlike [41], however, the focus here is not on the combination of handcrafted features and deep learning methods but rather on the implementation of powerful data augmentation approaches to strengthen the ensemble of CNNs. In this paper, different CNN topologies are trained using two different learning rates {0.001, 0.0001} [42], four different batch sizes {10, 30, 50, 70}, and multiple combinations of different data augmentation approaches, with two of them being proposed here for the first time. The new data augmentation approaches proposed here are *based* on two well-known feature transforms: the Discrete Cosine Transform (DCT) and Principal Component Analysis (PCA). Both methods are based on projecting the original image onto the DCT/PCA subspace, which perturbates the retro-projection from the subspace to the original space. Data augmentation via PCA has proposed in [43] and [44]. In [43], PCA is applied by perturbing the training images using extracted eigenvalues. In [44], the first principal component of each sample is multiplied by a random value drawn from a uniform distribution, then the images are back projected to the original space. In this work, we propose three different methods based on PCA.

It should also be noted that in [45] an ensemble of classifiers was proposed where the same network architecture was also repeated several times, with each repetition having different activation functions in the activation layers. In other words, each activation layer of each network was randomly substituted with a random activation selected from a pool.

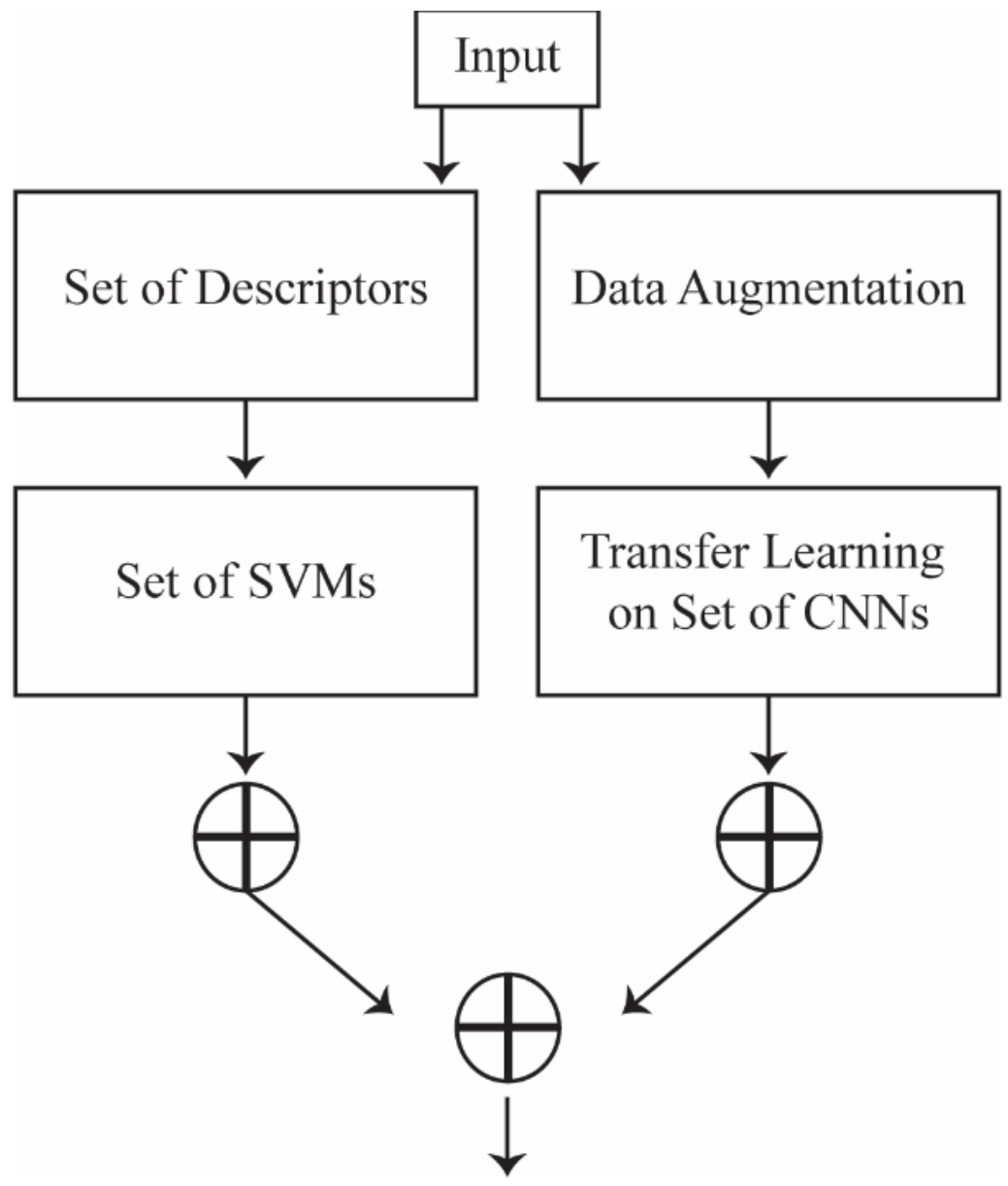

**Figure 1.** Schematic of the proposed ensemble system. The data augmentation step is performed only in the training step. In the test step, we use the fine-tuned CNNs.

To prove that our system is highly generalizable, we evaluate our method on a diverse set of bioimage classification problems, each represented by a benchmark dataset. Some of these datasets are publicly available in the IICBU 2008 database, and each bioimage task represents a typical subcellular, cellular, and tissue level classification problem. Results show that the proposed GenP bioimage ensemble obtains state-of-the-art performance without any ad hoc dataset-dependent tuning of parameters.

### *1.3. Organization of this Work*

This paper is organized as follows. The set of handcrafted methods evaluated in this work are listed in section 2. The CNN architectures used to build the ensemble of deep learners are presented in section 3. The new augmentations methods and stochastic ensembles are detailed in sections 3.1 and 3.2, respectively. Our experiments are described in section 4 along with a discussion of experimental results, and the study concludes in section 5 with suggestions for further research.

## 2. Handcrafted Methods

As noted above, our GenP bioimage ensemble combines both handcrafted and learned descriptors with some traditional and novel augmentation methods. The handcrafted descriptors considered in this work are described in Table 1, where we also provide the parameter settings used for each descriptor. In subsections 2.1-2.11, a detailed discussion of all these descriptors (except for MOR which is described in Table 1) is provided. Color images are processed by evaluating descriptors on the three channels separately and training an SVM for each one; the SVMs are then combined by sum rule as noted in Figure 1. The hyperparameters of the SVM are chosen using the training data.

**Table 1.** Summary of the handcrafted texture descriptors considered in this work.

| Acronym | Brief Description of Descriptor and Parameter Settings | Source |
|---|---|---|
| LTP | Multiscale Uniform Local Ternary Patterns, which is an extension of LBP [46] with two (*radius*, *neighboring points*) configurations: (1, 8) and (2, 16). | [47] |
| MLPQ | Multithreshold Local Phase Quantization is composed of sets of LPQ descriptors that vary the filter sizes, the scalar frequency, and the correlation coefficient between adjacent pixel values. Each extracted descriptor trains a different SVM. | [48] |
| CLBP | Completed LBP with two (*radius, neighboring points*) configurations: (1,8) and (2,16). | [49] |
| RIC | Multiscale Rotation Invariant Co-occurrence of Adjacent LBP with *radius* ∈ {1, 2, 4}. | [50] |
| GOLD | Gaussians of Local Descriptors. Here we train a different SVM from each region of the spatial pyramid and combine them by sum rule. We use one level spatial pyramid decomposition: the decomposition consists of the entire image, followed by level one, where the image is subdivided into four quadrants. | [51] |
| COL | A simple and compact set of soft color descriptors such as the mean and standard deviation of the color space (RGB, HSV, and CIE Lab). | [52] |
| AHP | Adaptive Hybrid Pattern combines *i)* a Hybrid Texture Model (HTD) composed of local primitive features and a global spatial structure, and *ii)* an adaptive quantization algorithm (AQA) to improve noise robustness. We fixed *quantization level* = 5; we used 2 (*radius, neighboring points*) configurations: (1, 8) and (2, 16). | [53] |
| FBSIF | Extension of the canonical Binarized Statistical Image Features (BSIF) by varying the parameters of filter size (SIZE_BSIF, *size* ∈ {3, 5, 7, 9, 11}) and a threshold (*th*) for binarizing (FULL_BSIF, *th* ∈ {-9, -6, -3, 0, 3, 6, 9}). | [54] |
| LET | A simple but effective representation, LET encodes the joint information within an image across feature and scale spaces. We use the default values available in the MATLAB toolbox. | [55] |
| CLM | CodebookLess Model. We use the ensemble called CLoVo_3 in [41] based on e-SFT, PCA for dimensionality reduction, and one-vs-all SVM for the training phase. | [56] |
| ETAS | We utilized Threshold Adjacency Statistics from a novel perspective to enhance discrimination power and efficiency. Seven threshold ranges were used to produce seven distinct feature spaces, which were then used to train a single SVM. The source code for this method is located here: https://drive.google.com/file/d/0B7IyGPObWbSqRTRMcXI2bG5CZWs/view. | [57] |
| MOR | Morphological descriptor is a set of measures (aspect ratio, number of objects, area, perimeter, eccentricity, etc.) extracted from a segmented version of the image as enumerated in [58]. | [58] |

### 2.1. Local Ternary Pattern (LTP)

LTP [47] extends LBP [46], which can be expressed as

$$\mathrm{LBP}_{\mathrm{P,R}} = \sum_{\mathrm{p}=0}^{\mathrm{P}-1} \mathrm{s}(x) 2^{\mathrm{P}}, \tag{1}$$

where $q_c$ is the pixel under consideration (the central point), and $q_p$ is a set of points surrounding $q_c$. This neighborhood is bounded by the radius R surrounding $q_c$ and the total number of pixels P included in the neighborhood. The variable $x = q_p - q_c$, or the difference between the intensity levels of $q_p$ and $q_c$. The function *s(x)* produces the binary encoding:

$$s(x) = \begin{cases} 1, x \geq 0 \\ 0, otherwise. \end{cases} \tag{2}$$

Each digit of LBP is either 0 or 1, with codes ranging in [0, $2^P$-1]. LBP descriptors are the histograms of these binary numbers.

The LTP extension of *s(x)* in [47] provides a ternary encoding using a threshold $\tau$ around zero:

$$s(x) = \begin{cases} 1, x \geq \tau \\ 0, -\tau \leq x < \tau \\ -1\ otherwise. \end{cases} \tag{3}$$

The resulting ternary codes are much larger than LBP. For this reason, LTP histograms are split into binary sub-histograms that are concatenated: each LTP code is divided into a positive and negative binary pattern according to the sign of its components.

### 2.2. Multi-threshold Local Phase Quantization with ternary coding (MLPQ)

MLPQ [48] extends LPQ by using a multi-threshold approach, with threshold values $\tau \in \{0.2, 0.4, 0.6, 0.8, 1\}$. In this study we follow the protocol in Nanni et al. [41] that combines sets of LPQ extracted with filter size $\mathrm{R} \in \{1, 3, 5\}$ and correlation coefficient between adjacent pixel values $\mathrm{P} \in \{0.75, 0.95, 1.15, 1.35, 1.55, 1.75, 1.95\}$). Every descriptor that is extracted is fed into a separate SVM.

### 2.3. Completed Local Binary Pattern (CLBP)

CLBP [49] is so named because this descriptor represents a local region using the region's center pixel (CLBP-C) and the Local Difference Sign-Magnitude Transform (LMSMT). This transform calculates two components: the difference signs, called CLBP-Sign (CLBP_S), and the difference magnitudes, called CLBP-Magnitude (CLBP_M). CLBP is the histogram formed by the three binary representations of CLBP_C, CLBP_S, and CLBP_M.

More formally, given a central pixel $q_c$ along with its P neighbors and letting $d_p = q_c - q_p$, then $d_p$ can be decomposed into the two LMSMT components, *CLBP_S* and *CLBP_M*, as follows:

$$d_p = S_p * m_p \text{ and } \begin{cases} S_p = sign\ (d_p) \\ m_p = |d_p| \end{cases}$$

$$S_p = \begin{cases} 1, d_p \geq 0 \\ -1, d_p < 0, \end{cases} \tag{4}$$

where $S_p$ represents the sign of $d_p$, and $m_p$ the magnitude of $d_p$. As can be observed, LDSMT transforms the vector $[d_0, \ldots, d_{P-1}]$ into both the sign vector $[s_0, \ldots, s_{P-1}]$ and the magnitude vector $[m_0, \ldots, m_{P-1}]$.

CLBP_S can now be defined similarly to LBP as in Eq. (1):

$$CLBP_{M_{P,R}} = \sum_{p=0}^{P-1} t(m_p, c) 2^P,\ t(x, c) = \begin{cases} 1, x \geq c \\ 0, x < c. \end{cases} \tag{5}$$

where c is the mean value of *m*.

### 2.4. Rotation Invariant Co-occurrence among adjacent Local Binary Pattern (RIC)

The main idea of RIC [50] is simple. LBP fails to preserve structural information in the binary patterns. To retain this information, RIC represents using LBP pairs. Letting $I$ be an image and $r = (x, y)$ a position vector in $I$, LBP can also be expressed as

$$LBP(\boldsymbol{r}) = \sum_{p=0}^{P-1} s(I(\boldsymbol{r} + \Delta_{d_p}) - I(\boldsymbol{r}))2^p, \tag{6}$$

where P is the number of neighbor pixels and $\Delta_{d_p}$ is the displacement vector from the center pixel to the neighboring pixels given by $\Delta_{d_p} = (d\cos(\theta_p), d\sin(\theta_p))$, where $\theta_p = 2\pi/P_p$ and $s(x)$ is the scaled parameter of LBP in Eq.(2).

The co-occurrence among LBPs over an image can be converted into LBP pairs at a given pixel $r$ thus:

$$Pair(\boldsymbol{r}, \Delta\boldsymbol{r}) = (LBP(\boldsymbol{r}), LBP(\boldsymbol{r} + \Delta\boldsymbol{r})), \tag{7}$$

where $\Delta\boldsymbol{r} = (r\cos\theta, r\sin\theta)$ is a displacement vector between an LBP pair, with $r$ an interval between an LBP pair, and $\theta = 0, \pi/4, \pi/2, 3\pi/4$. Rotation invariance can be incorporated by assigning a rotation invariant label to each LBP pair equal in rotation.

### 2.5. Gaussians of Local Descriptors (GOLD)

GOLD [51] improves the canonical Bag of Words (BoW), which extracts local features in order to create a codebook from which a global image representation can be produced. GOLD is a four-step process:

1. Dense SIFT descriptors are extracted on a regular grid of the input image.
2. The image is decomposed into regions by a multilevel recursive image decomposition and softly assigned to regions according to a local weighting.
3. Each region is represented as a multivariate Gaussian distribution of the extracted local descriptors via a local mean and covariance inference.
4. The covariance matrix is projected on the tangent space and concatenated to the mean to obtain the final region descriptor.

Descriptors are vectorized by projecting the covariance matrix on a Euclidean space and concatenating the mean vector.

### 2.6. Color Descriptors (COL)

To produce the COL [52] descriptors,The following soft descriptors of the color space (RGB, HSV, and CIE Lab) are computed in terms of the mean, standard deviation, and $k$-th moment, defined as

- *Mean*: $\mu_c = \frac{1}{n}\sum_{i=1}^{n} I_{c,i}I;$ (8)
- *Standard deviation*: $\sigma_c = \frac{1}{n-1}\sqrt{\sum_{i=1}^{n}(I_{c,i} - \mu_c)^2};$ (9)
- *k-th moment*: $m_c = \sum_{i=1}^{n}(I_{c,i} - \mu_c)^k h_c(I_{c,i});$ (10)

  where $n$ is the number of pixels in the image, $I_{c,i} \in [0,1]$ is the intensity of the $i$-th pixel in the $c$-th color channel, and $m_c$ is the average intensity value of the $c$-th color channel.

### 2.7. Adaptive Hybrid Pattern (AHP)

AHP [53] uses a Hybrid Texture Model (HTD) to overcome problems with LBP's sensitivity to noise. HTD is composed of local primitive features and global spatial structures. HTD is then combined with an adaptive quantization algorithm (AQA) to improve the noise robustness of the angular space quantization since the vector quantization thresholds are adjusted based on the content of the local patch.

Let $X$ be an input random variable for quantizing, then the variation interval of $X$ is A. Now define partition $S = \{S_i; i \in I\}$ as $S_i \cap S_j = \emptyset$, $i \neq j$, $\cup_{i \in x} S_i = A$, where $i$ is the index of the subset and $I = \{0, 1, 2, \ldots, n-1\}$, with $n$ the number of subsets. The boundaries or thresholds of each subset are defined as $S_i = [Thr_i, Thr_{i+1}]$, where $\forall i \in I, Thr_i < Thr_{i+1}$.

The quantization function can be defined as $s': A \rightarrow Z$, where $Z = I = \{0, 1, 2, \ldots, n-1\}$ and $n$ is the quantization level. The boundaries of all $S_i$ are the solutions of

$$\int_{Thr_i}^{Thr_{i+1}} f_x(X) = dX = \frac{1}{n}, \qquad \forall i \in Z, \tag{11}$$

where $f_x$ is the probability density function describing the input variable $X$.

The quantization thresholds for $T_{global}$ are set as

$$Thr_G_i = \sqrt{2}erf^{-1}(2i - n/n) \cdot \sigma_I, i = 1,2,\ldots, n-1, \tag{12}$$

where $\mathrm{erf}\,(\cdot)$ is the error function, and $\sigma_I$ is the standard deviation of the mean value of the whole image texture.

Letting $\sigma_L$ be the standard deviation of $q_p$ and given Eq. (8), the quantization thresholds for $T_{local}$ are

$$Thr_L_i^s = \begin{cases} \frac{\sqrt{2}}{2}\ln\frac{2i}{n} \cdot \sigma_s & i < n/2 \\ -\frac{\sqrt{2}}{2}\mathrm{n}\frac{2n-2i}{n} \cdot \sigma_s & i < n/2 \end{cases}, \; s = L, G, \tag{13}$$

where L and G indicate *local* and *global*, respectively.

The length of the histogram can be compressed by splitting the global pattern and the local pattern into many binary patterns.

The algorithm for computing AHP features is a five-step process:

1. Compute the thresholds for adaptive quantization using Eq. (12) and (13).
2. For each local patch with radius R and neighborhood P, compute the local patterns and global patterns as follows:
3. $AHP_G_i = \sum_{p=0}^{P} s(q_p - q_I - Thr_G_i)\, 2^P$, i=1,…,n-1,
4. $AHP_L_j^s = \sum_{p=0}^{P} s(q_p - q_c - Thr_L_j^s)\, 2^P$, j=1,…,n-1,
5. where $s = L, G$.
6. Map each binary pattern into rotation invariant uniform (*riu*2) or uniform (*u*2) patterns;
7. Compute the statistical histogram of $AHP_G_i$ and $HP_L_j^s$ accumulated over the whole texture image;
8. Output the joint combination of the histograms from step 4 as the final AHP texture descriptor.

### *2.8. Full Binarized Statistical Image Features (FBSIF)*

FBSIF [54] is an extension of the Binarized Statistical Image Features (BSIF). With a set (*n*) of linear filters, BSIF assigns an *n*-bit label to each pixel in a given image. These assignments project local image patches (size of $l\,x\,l$ pixels) onto a subspace.

The *n*-bit label is represented in binary form as

$$s = \boldsymbol{WX}, \tag{14}$$

where $\boldsymbol{W}$ is a $n \times l^2$ matrix containing the compilation of the vector notations of the filters and **X** is the $l^2 \times 1$ vector notation of the $l \times l$ neighborhood. If the vector $\boldsymbol{x}$ is defined as containing the pixels of the $l \times l$ neighborhood, and $\boldsymbol{w}_i$ is the *i*-th row of **W**, then the *i*-th digit of $s$ is a function of the *i*-th linear filter $\boldsymbol{w}_i$ expressed as

$$s_i = \boldsymbol{w}_i^T \boldsymbol{x}. \tag{15}$$

with each bit of the BSIF code defined as

$$b_i = \begin{cases} 1, if\ s_i > 0 \\ 0, if\ s_i \leq 0 \end{cases} \tag{16}$$

The set of filters $\boldsymbol{w}_i$ is produced by maximizing the statistical independence of the filter responses $s_i$ on a set of patches in an image by applying Independent Component Analysis (ICA).

FBSIF is an extension of BSIF. FBSIF alters the filter size (SIZE_BSIF, *size* ∈ {3, 5, 7, 9, 11}) and a threshold *th* used for binarizing (FULL_BSIF, *th* ∈ {-9, -6, -3, 0, 3, 6, 9}). This process produces thirty-five (*size*, *th*) combinations, each of which trains a separate SVM. The set of SVMs is then combined via the sum rule.

### 2.9. Locally Encoded Transform feature histogram (LET)

LET [55] is a histogram that encodes the joint information within an image across feature and scale spaces. Computing LET descriptors is a three-step process:

1. Transform features that characterize local texture structures and their correlation are constructed (this is accomplished by applying linear and non-linear operators on the responses of directional Gaussian derivative filters in scale space).
2. Scalar quantization is performed with binary or multilevel thresholding used to quantize the transform features into texture codes.
3. The discrete texture codes are aggregated into a histogram representation.

### 2.10. Codebookless Model (CLM)

CLM [56] is a variant of the Bag of Feature (BoF) [59]. CLM substitutes the dictionary in BoF with a single Gaussian. $N$ local features $\{\boldsymbol{x_i} \in \mathbb{R}^{k\times 1}, i = 1, \dots, N\}$ are extracted from a given image on a dense grid using the maximum likelihood method, and the features are represented with a gaussian model as:

$$(x_i|\mu, \Sigma) = \frac{exp\left(-\frac{1}{2}(x_i-\mu)^T\Sigma^{-1}(x_i-\mu)\right)}{\sqrt{(2\pi)^k \det(\Sigma)}}, \quad (17)$$

where det(·) is the matrix determinant, $\mu = {}^1/_n \sum_{i=1}^{N} x_i$ is the man vector, and $\Sigma = \sum_{i=1}^{N}(x_i-\mu)(x_i-\mu)^T$ is the covariance matrix.

Multiscale SIFT descriptors [60] are then extracted. This is accomplished with the standard BoF pipeline having a cell size $2^i$, $i = \{1,2,\dots,n\}$ and with a single scale pixel-wise covariance descriptor [61] using dense sampling (step-length = 2). The dense covariance descriptors are calculated using a raw feature vector of length seventeen. This vector provides information about the intensity of the image and four kinds of first-order and second-order gradients. Matrix logarithm is evaluated on the covariance descriptors (LogCov) and vectorized. In this work, we use the VLFeat library [62] to calculate the SIFT features.

Concatenated with the SIFT and LogCov descriptors is more information about the gradient, scale, entropy, color, and location of the image. The width and height of the image are also resized to a minimum of sixty-four pixels to ensure sufficient data for calculating the covariance matrices and Gaussian models. It is also necessary to add $\varepsilon=10^{-3}$ to avoid zeros in the diagonal of the covariance matrices. The image is further divided into regular regions (such as $1 \times 1$, $2 \times 2$, $3 \times 3$, and $4 \times 4$) using a spatial pyramid strategy. A Gaussian model is computed on each region and concatenated to represent the whole image.

### 2.11. Extended Threshold Adjacency Statistics (ETAS)

ETAS [57] utilizes Threshold Adjacency Statistics and seven threshold ranges to generate seven feature sets that are fed into seven separate SVMs. A given input image is binarized using the following seven threshold values $E_x$:

$$\left.\begin{array}{c} E_1 = \mu \; to \; 255 \\ E_2 = \mu - \tau \; to \; 255 \\ E_3 = \mu - \tau \; to \; \mu + \tau \\ E_4 = \mu \; to \; 255 - \tau \\ E_5 = \mu - \tau \; to \; 255 - \tau \\ E_6 = \mu + \tau \; to \; 255 - \tau \\ E_7 = \mu + \tau \; to \; 255 \end{array}\right\} \text{Threshold Ranges}, \quad (18)$$

where $\mu$ is the average intensity of the image, and $\tau$ is the limit or cut-off value for calculating the average. The seven resulting binary images are then used to extract nine statistical features.

## 3. Deep Learner Ensembles and Augmentation

CNNs are a class of deep feed-forward regularized multilayer perceptrons that are scale, shift, and distortion invariant and composed of interconnected neurons arranged in three dimensions (width, height, and depth). Every layer in a CNN transforms a 3D input volume into a 3D output volume of neuronal activations. The basic CNN is composed of five layers: convolutional (CONV), activation (ACT), pooling (POOL), Fully-Connected (FC), and classification (CLASS). As noted in the Introduction, CNNs automatically extract features during the training process. In general,

CONV layers are responsible for feature extraction and are processed sequentially. The ACT layer is used for thresholding each input element using, e.g., the tanh or ReLu function. The POOL layer takes the output of the ACT layer and performs a downsampling according to two parameters: pool size (the rectangular size considered for downsampling) and stride value (the step size for processing the rectangular window). One or more FC layer are commonly placed at the end of the network, just before the final CLASS layer.

In this work, we evaluate ensembles composed of the following classic CNN architectures pretrained on the ImageNet database:

- AlexNet [27]: the first GPU-implementation of a CNN and winner of the 2012 ImageNet Large Scale Visual Recognition Challenge. This architecture uses the Rectified Linear Unit (ReLU) function and overlapping POOL layers;
- GoogleNet [63]: a CNN architecture that includes an inception module that approximates a sparse CNN with a normal dense construction. This architecture has two POOL and two CONV layers along with nine inception layers, each one consisting of six CONV and one POOL layer;
- VGGNet16 & VGGNet19 [64]: architectures from the VGG group that improves AlexNet by replacing large kernel-sized filters with multiple 3×3 kernel-sized filters;
- ResNet50 [65]: a 50-layers dee CNN network available in MATLAB;
- DenseNet: [66]: a logical extension of ResNet that connects each layer to every other layer;

In our experiments, each CNN is fine-tuned on each of the tested datasets. Fine-tuning a CNN is a procedure that essentially resumes the training process of a given pre-trained network so that it learns a new classification problem. The net is therefore initialized using pre-trained weights obtained on the large ImageNet dataset and further trained using the training set of the target problem. All the networks are modified by changing the last FC and CLASS layers to fit the number of classes of the target problem, without freezing the weights of the previous layers.

A CNN that fails to converge is excluded from the ensemble. We also exclude CNNs trained with large batch sizes or that result in a "GPU out of memory" error.

### *3.1. Augmentation Methods*

It was noted in the introduction that Data Augmentation (DA) is not only a highly effective way to enlarge a small training set but also a method for enhancing performance and reducing overfitting during training.

Our DA workflow can be described as follows. At the beginning of each epoch, we randomly transform each image in a given dataset with some basic preprocessing methods, such as rotation and reflection. Following this random preprocessing stage, four different data augmentation protocols (App1-4) are applied. These augmentations are performed on the fly on every batch. Hence, they do not require more memory, but they do increase the training time.

The four DA protocols used in this study are the following:

1. App1: The image is reflected in the left-right direction with 50% probability.
2. App2: The image is randomly reflected in both the left-right and the top-bottom directions. In addition, App2 linearly scales the image along both axes by two different factors that are randomly sampled from the uniform distribution in [1, 2].
3. App3: Combines all the transformations in App2 and adds image rotation and translation in both directions. The rotation is accomplished using an angle that is randomly sampled from the interval [-10, 10], while the translation consists of shifting the image by a certain number of pixels randomly sampled from [0, 5].
4. App4: Extends App3 by also applying vertical and horizontal shear, with the shear angles randomly sampled from the interval [0, 30].
   In each of these protocols (App1-App4), data augmentation randomly perturbs the training data for each epoch, so that each epoch uses a slightly different data set. The actual number of training images at each epoch does not change.

In addition to the four different DA protocols, we applied two new approaches presented for the first time here that are based on two common feature transforms: DCT and PCA. Both these transforms are based on the projection of the original image onto the DCT/PCA subspace and perturb the backprojection from the subspace to the original space.

Our two new DA approaches are labeled as follows:

1. App5: DA approach based on PCA;
2. App6: DA approach based on DCT.

PCA [67] is a popular method for image compression, and it is often used as an unsupervised dimensionality reduction method. Computationally cheaper to compute than PCA, DCT maps feature vectors into a smaller number of uncorrelated directions calculated to preserve the global Euclidean structure. Similar to DCT, PCA also extracts an orthogonal projection matrix by maximizing the variance of the projected vectors. DCT provides a good compromise between information packing and computational complexity [68]. Computational complexity is reduced because DCT is not data-dependent, while PCA is based on the eigenvalue decomposition of the data covariance matrix. DCT components are also small in magnitude since most salient information exists in the coefficients with low frequencies. However, discarding the transform coefficients corresponding to the highest frequencies from the representation produces small errors in image reconstruction.

Once the PCA and DCT coefficients of the decompositions are calculated, we propose three different methods for generating new images. In the first method, every component of the feature vector obtained with PCA/DCT is randomly set to zero with a given probability. Then, the inverse of the transform is performed on the new feature vector, and a new image is generated. In the second method, some of the features extracted in random points selected using a gaussian distribution are reset. After that, the inverse of the transform is performed. In the third method, five random images in the dataset are selected that have the same label as a given image. We then perform a feature transform on all six images to obtain their feature vectors via the PCA/DCT transform. At this point, we randomly exchange some of the features of the original image with some of the corresponding features of the five randomly selected images. We then perform the inverse of the transform to generate the new image, which is a mixture of the six images, and label it the same as the others the augmented image was generated from. We do this for each image of the training set adding the new images to it.

Below is the pseudocode for the new DA methods. For the sake of brevity, we report the pseudocode for DCT only. The methods based on PCA are the same, except that the PCA space is built using the training data instead of the DTC space. The images generated by PCA and DCT are also reflected in the left-right direction with 50% probability for further augmentation. Notice that for each image we have three channels (RGB).

*Pseudocode of the three DCT-based data augmentation approaches.*

**Algorithm** *ResetCoef*

**Input**: Image: tensor $n \; x \; n \; x \; 3$

**Output**: NewImage: tensor $n \; x \; n \; x \; 3$

channel ← 1;

**for** every channel **do**

 *#DCTimage is a matrix of dimension* $n \; x \; n$

 DCTimage ← calculateDCT(Image(:, :, channel)); *# see Eq. 1*

 **for** row,col in DCTimage **do**

  with probability 0.5 **do**

   DCTimage(row, col) = 0; #except DCTimage(1,1) that cannot be reset

  **end**

 **end**

 *#inverse of the perturbated image*

 NewImage(:, :, channel) ← inverseDCT(DCTimage);

**end**

**Algorithm** *Perturbation*

**Input**: Image, tensor $n \; x \; n \; x \; 3$

**Output**: NewImage, tensor $n \; x \; n \; x \; 3$

channel ← 1;

**for** every channel **do**

*#DCTimage is a matrix of dimension* $n\ x\ n$

DCTimage ← calculateDCT(Image(:, :, channel)); *# see Eq. 1*

Sigma = standardDeviation(Image)/2;

**for** row,col in DCTimage **do**

DCTimage(row, col) += sigma × random number $z \sim U\left(-\frac{1}{2}, \frac{1}{2}\right)$;

# except for DCTimage(1,1), which cannot be modified.

**end**

*#inverse of the perturbated image*

NewImage(:, :, channel) ← inverseDCT(DCTimage);

**end**

<u>**Algorithm** *FusionImg*</u>

**Input**:Image: tensor $n\ x\ n\ x\ 3$

Images : list of $n\ x\ n\ x\ 3$ tensors

**Output**: tensor NewImage, $n\ x\ n\ x\ 3$

sample1,…,sample5 ← random images in Images whose label is the same of image

channel ← 1;

**for** every channel **do**

*#DCTimage is a matrix of dimension n x n*

DCTimage ← calculateDCT(Image(:, :, channel)); *# see Eq. 1*

**for** sample = sample1,…,sample5 **do**

sampleDCT = calculateDCT(sample(:, :, channel));

**for** row,col in DCTimage **do**

with probability 0.05 **do**

DCTimage(row, col) = sampleDCT(row,col);

**end**

**end**

**end**

*#inverse of the perturbated image*

NewImage(:, :, channel) ← inverseDCT(DCTimage);

**end**

The formula for calculating DCT used in the three algorithms presented in above is the following:

$$DCTimage(x,y) = \sum_{p,q=1}^{n} a_p a_q Image(p,q) \cos\frac{2p-1}{2n} \cos\frac{2q-1}{2n}, \quad (18)$$

where $a_p = \begin{cases} \sqrt{\frac{1}{n}},\ n = 1 \\ \sqrt{\frac{2}{n}},\ n > 1 \end{cases}.$

*3.2. Stochastic approach*

The stochastic approach creates an ensemble of networks starting from a given architecture and a pool of different activation functions. We create a new random network by randomly substituting all the activations in the original architecture with a new one among those in the pool. Note that a different activation is selected in every layer. This algorithm can be repeated multiple times and yield a different network each time. Hence, these networks can be grouped together to create an ensemble. In our experiments, we use ResNet50 as our backbone architecture and a large pool of activations. The functions that we include in the pool are ReLU, leaky ReLU, ELU, PReLU, S-Shaped ReLU, Adaptive Piecewise Linear Unit, MeLU, GaLU, Symmetric MeLU, Symmetric GaLU, flexible MeLU, MeLU+GaLU, Tanh+ReLU, Splash, and Parametric Deformable Linear Unit [69].

## 4. Experimental Results

### *4.1. Datasets*

Several datasets that include very different image types were selected to test our system and demonstrate the generalizability of our GenP bioimage system. In order to ease the comparison of other systems with the one proposed here, the datasets used in our experiments are all publicly available:

- CH: the CHINESE HAMSTER OVARY CELLS [70] dataset of 327 fluorescent microscopy images, size 512×382, that are divided into five classes. This dataset is located at http://ome.grc.nia.nih.gov/iicbu2008/hela/index.html#cho;
- HE: the 2D HELA dataset [70] of 862 images, size 512×382, of HeLa cells acquired by fluorescence microscope and divided into ten classes. This dataset is located at http://ome.grc.nia.nih.gov/iicbu2008/hela/index.html;
- LO: the LOCATE ENDOGENOUS [71] dataset of 502 images, size 768×512, of mouse sub-cellular images showing endogenous proteins or specific organelle features. The images are unevenly divided into ten classes. This dataset is located at http://locate.imb.uq.edu.au/downloads.shtml;
- TR: the LOCATE TRANSFECTED dataset of 553 mouse sub-cellular images, size 768×512, showing fluorescence-tagged or epitope-tagged proteins transiently expressed in specific organelles [71]. The images are unevenly divided into 11 classes. This dataset is located at http://locate.imb.uq.edu.au/downloads.shtml;
- RN: the FLY CELL dataset [71] of 200 images, size 1024×1024, of fly cells acquired by fluorescence microscopy and divided into ten classes. This dataset is located at http://ome.grc.nia.nih.gov/iicbu2008/rnai/index.html;
- MA: Muscle aging [72] dataset of images, size 1600×1200, of C. elegans muscles at four ages. This dataset is located at https://ome.grc.nia.nih.gov/iicbu2008;
- TB: Terminal bulb aging [72] dataset of images, size 300×300, of C. elegans terminal bulb at seven ages (hence, seven classes). This dataset is located at https://ome.grc.nia.nih.gov/iicbu2008;
- LY: Lymphoma [72] dataset of malignant lymphoma images, size 1388×1040, of three subtypes. This dataset is located at https://ome.grc.nia.nih.gov/iicbu2008;
- LG: Liver gender [72] dataset of images, size 1388×1040, of liver tissue sections from 6-month male and female mice on a caloric restriction diet (the classes are the two genders). This dataset is located at https://ome.grc.nia.nih.gov/iicbu2008;
- LA: Liver aging [72] dataset of images, size 1388×1040, of liver tissue sections from female mice on an ad-libitum diet of 4 ages. This dataset is located at https://ome.grc.nia.nih.gov/iicbu2008;
- CO: Collection of images, size 150×150, of textures in histological images of human colorectal cancer [73]. This dataset is located at https://zenodo.org/record/53169#.WaXjW8hJaUm.

Unless specified otherwise in the description of the datasets above, the protocol used in our experiments was the five-fold cross-validation method. To avoid overfitting, the same set of descriptor parameters were used for all descriptors across all tested datasets. The following experiments were statistically validated using the Wilcoxon signed-rank test [74].

### *4.1. Experiments*

In Table 2, we report the performance obtained by some of the handcrafted features and the following ensembles:

- FH: sum rule among LTP, MLPQ, CLBP, RICLBP, LET, MOR, AHP, FBSIF, COL (on only the datasets with colored bioimages) and ETAS;
- FH-etas: same as FH but without considering ETAS;
- FUS1: sum rule of FH and CLM
- FUS2: sum rule of FUS1 and GOLD.

Before each fusion, the scores of the SVMs trained with a given descriptor are normalized to mean 0 and standard deviation 1.

In the last row of Table 2, we report the performance of the handcrafted ensembles tested in Nanni et al. [41], labeled PREVIOUS. In the column labeled Avg, we report the average accuracy obtained by a given descriptor/ensemble across the entire set of tested datasets.

Examining Table 2, we find that FBSIF and MLPQ obtained the best performances among the tested individual descriptors. There is no statistical difference between these two methods, however. Both outperform all the other individually tested methods with a p-value of 0.1. The best performing ensembles are FUS1 and FUS2. They outperform the other ensembles, as well as the best performing individual descriptors FBSIF and MLPQ, with a p-value of 0.1.

In Table 3, we compare the different approaches for data augmentation, reporting, for the sake of space, the performances using ResNet50 and DenseNet. The method labeled ENS is the sum rule among the CNNs trained using the six data augmentation approaches. The performance reported in Table 3 is the sum rule of each method trained with the two different learning rates ({0.001, 0.0001}) and different batch sizes ({10, 30, 50, 70}), as reported in the previous section. The method named E_ReLu is an ensemble of 15 CNNs all trained with a learning rate of 0.001 and a batch size of 30. E_ReLu represents a baseline for comparing the ensembles built considering different data augmentation approaches. SA is the performance of a single CNN trained with a learning rate of 0.001 and a batch size of 30 (which, on average, is the optimum set of hyperparameters).

**Table 2.** Performance accuracy of local-based approaches and their fusion across the considered datasets

| | CH | HE | LO | TR | RN | TB | LY | MA | LG | LA | CO | Avg |
|---|---|---|---|---|---|---|---|---|---|---|---|---|
| LTP | 98.77 | 87.33 | 94.6 | 90.55 | 80 | 55.88 | 85.33 | 78.75 | 98.00 | 98.67 | 90.40 | 87.11 |
| MLPQ | 99.38 | 92.79 | 97.6 | 97.09 | 88.5 | 62.89 | 92.27 | 91.67 | 99.33 | 99.81 | 93.58 | 92.26 |
| CLBP | 94.15 | 89.42 | 86.2 | 84 | 70 | 61.03 | 86.67 | 75.42 | 96.00 | 99.24 | 92.04 | 84.92 |
| RICLBP | 96.62 | 85.35 | 92.6 | 91.82 | 82 | 54.54 | 85.87 | 91.67 | 99.33 | 99.62 | 91.56 | 88.27 |
| LET | 97.85 | 92.33 | 95.80 | 92.91 | 75.00 | 54.85 | 92.53 | **98.75** | **100** | 99.81 | 93.18 | 90.27 |
| MOR | 97.85 | 84.88 | 93.60 | 92.36 | 83.50 | 56.60 | 84.53 | 80.00 | 96.33 | 98.29 | 93.30 | 87.38 |
| GOLD | 92.62 | 85.81 | 87.8 | 75.45 | 50 | 55.05 | 53.07 | 66.67 | 85.00 | 39.24 | 83.58 | 70.39 |
| AHP | 98.77 | 91.86 | 96.4 | 95.45 | 88 | 59.48 | 93.87 | 90.42 | 98.67 | 99.81 | 94.16 | 91.53 |
| FBSIF | 99.38 | 94.19 | 98.2 | 98.55 | 87 | 65.67 | 92.53 | 88.75 | **100** | 99.81 | 93.42 | 92.50 |
| COL | --- | --- | --- | --- | --- | --- | 91.47 | --- | 99.67 | 99.62 | 92.30 | --- |
| ETAS | 84.92 | 73.95 | 95.00 | 84.91 | 59.50 | 51.03 | 87.73 | 69.58 | 98 | 98.29 | 92.04 | 81.35 |
| CLM | 98.15 | 91.05 | 95.40 | 90.73 | 82.00 | 68.56 | 74.40 | 91.67 | 99.67 | 96.95 | 89.60 | 88.92 |
| FH | 99.69 | 93.95 | 98.60 | 98.55 | 91.00 | 68.35 | 94.67 | 92.08 | **100** | **100** | **95.20** | 93.82 |
| FH-etas | 99.69 | 93.95 | 98.20 | 98.55 | 90.50 | 68.04 | 94.13 | 91.67 | **100** | **100** | 95.18 | 93.62 |
| FUS1 | **100** | 94.88 | **98.80** | **98.91** | **92.00** | 71.24 | **94.67** | 92.50 | **100** | **100** | 95.18 | 94.38 |
| FUS2 | **100** | **95.70** | **98.80** | 98.36 | **92.00** | **71.86** | 93.87 | 93.33 | **100** | **100** | 94.94 | **94.44** |
| PREVIOUS | 99.69 | 94.42 | 98.40 | 98.36 | 90.50 | 70.62 | 92.00 | 91.67 | **100** | 99.62 | 93.74 | 93.54 |

Clearly, the best approach is given by ENS, which outperforms all the other approaches with a p-value of 0.1. Among the stand-alone methods, the best performance is obtained by the DCT-based method. These results demonstrate the value in using feature transforms for augmenting datasets and improving the performance of CNN.

Finally, in Table 4 we compare the performance of some of our ensembles with several state-of-the-art approaches reported in the literature. The following ensembles are reported in Table 4:

- HAND: the method named FUS1 in Table 2;
- DEEP: an ensemble of all the trained CNNs (using different the different values for LR, BS, and DA);

- DEEP(1-4): the same as DEEP but using only four CNNs (AlexNet; GoogleNet, VGGNet16, and VGGNet19);
- STOC: the stochastic ensemble detailed in section 3.2 (CO is not performed due to computation time);
- E_RELU - DENSENET: an ensemble of 15 DenseNet all trained with learning rate 0.001 and batch size 30;
- HAND+DEEP: sum rule between HAND and DEEP.

When we combine two methods by sum rule, their scores are normalized before the fusion to mean 0 and standard deviation 1.

**Table 3.** Performance accuracy of different configurations for data augmentation. Each App*X* is defined in section 3.1, and all the CNNs are trained using standard ReLU.

| | Param Set | CH | HE | LO | TR | RN | TB | LY | MA | LG | LA | CO | Avg |
|---|---|---|---|---|---|---|---|---|---|---|---|---|---|
| Resnet50 | SA | 95.86 | 92.98 | 95.65 | 94.63 | 57.57 | 62.71 | 85.30 | 88.42 | 96.49 | 93.02 | 95.58 | 87.11 |
| | App1 | 98.15 | 94.42 | 98.40 | 96.55 | 81.00 | 70.41 | 87.73 | **98.33** | 98.33 | 96.38 | 95.30 | 92.27 |
| | App2 | 98.15 | 94.30 | 97.80 | 96.00 | 65.00 | 68.66 | 87.73 | 93.75 | 99.33 | 94.86 | 96.72 | 90.20 |
| | App3 | 96.62 | 93.14 | 97.20 | 96.55 | 60.00 | 67.84 | 89.87 | 90.00 | 99.67 | 94.48 | 96.46 | 89.25 |
| | App4 | 98.15 | 95.58 | 97.00 | 96.00 | 64.00 | 67.94 | 88.00 | 83.75 | **100** | 97.14 | 96.40 | 89.45 |
| | App5 | 97.54 | 94.42 | 98.80 | 96.55 | 72.50 | 67.11 | 85.87 | 96.67 | 96.67 | 92.00 | 96.72 | 90.44 |
| | App6 | 98.77 | 95.93 | **99.00** | 98.00 | **83.00** | 72.47 | 89.33 | 96.67 | 98.67 | 98.10 | 96.46 | 93.30 |
| | ENS | **99.38** | 95.00 | **99.00** | 98.00 | 82.50 | **73.81** | 91.20 | 97.92 | 99.33 | **98.86** | **97.40** | **93.85** |
| | E_ReLu | 99.08 | **96.05** | 98.40 | **98.55** | 66.00 | 70.21 | **92.27** | 95.83 | 99.33 | 98.67 | **97.40** | 91.98 |
| Densenet | SA | 97.87 | 95.16 | 97.07 | 95.76 | 64.17 | 64.32 | 85.71 | 88.97 | 94.00 | 93.57 | 95.73 | 88.39 |
| | App1 | **99.69** | 96.28 | 98.40 | 97.82 | 81.00 | 70.62 | 88.00 | 95.83 | 99.33 | 97.52 | 95.72 | 92.74 |
| | App2 | 98.77 | 95.58 | 98.20 | 97.27 | 74.00 | 71.55 | 91.47 | 91.25 | 80.67 | 95.05 | 96.32 | 90.01 |
| | App3 | 98.46 | 96.16 | 97.80 | 96.55 | 74.00 | 67.63 | 90.13 | 90.42 | 99.67 | 99.05 | 96.46 | 91.48 |
| | App4 | 98.46 | 95.93 | 98.00 | 97.27 | 71.00 | 70.72 | 89.07 | 92.50 | 87.67 | 97.71 | 96.72 | 90.45 |
| | App5 | **99.69** | 96.28 | 98.40 | 97.64 | 78.50 | 71.86 | 86.13 | 95.42 | 98.67 | 97.52 | 97.14 | 92.47 |
| | App6 | **99.69** | 96.28 | **99.20** | **98.18** | 81.00 | 74.64 | 88.27 | **97.92** | **100** | 98.86 | 97.14 | 93.74 |
| | ENS | **99.69** | 96.28 | 98.80 | **98.18** | **84.50** | 74.02 | 92.53 | 95.83 | **100** | **99.62** | 97.26 | **94.24** |
| | E_ReLu | 99.38 | **96.40** | 98.40 | 98.55 | 79.00 | 71.24 | 86.40 | 94.58 | 99.67 | 99.24 | **97.84** | 92.79 |

HAND + DEEP is the best ensemble proposed here: it outperforms all the other approaches reported in the experimental section with a p-value of 0.1. Clearly, the proposed ensemble outperforms the ensembles in [41] and [75]. It should be noted that we tried to improve the performance of HAND + DEEP by adding Stoc to it, but the performance was very similar.

Unlike the other state-of-the-art methods, the MATLAB source code for reproducing results is freely available. Given that all the descriptors can be calculated in parallel by exploiting the modern multicore CPUs (for handcrafted features) and GPUs (for deep learning features), all the descriptors can be extracted in a reasonable amount of time for all applications where real-time computation is not important (which is the case for many medical image classification problems).

**Table 4**. Comparison with state of the art with accuracy as the performance indicator.

| | CH | HE | LO | TR | RN | TB | LY | MA | LG | LA | CO |
|---|---|---|---|---|---|---|---|---|---|---|---|
| Hand | **100.00** | 94.88 | 98.80 | 98.91 | 92.00 | 71.24 | 94.67 | 92.50 | **100.00** | **100.00** | 95.18 |
| E_ReLu - Densenet | 99.38 | 96.40 | 98.40 | 98.55 | 79.00 | 71.24 | 86.40 | 94.58 | 99.67 | 99.24 | 97.84 |
| Deep | 99.38 | 96.51 | **99.20** | 98.55 | 86.50 | 74.64 | 92.80 | 98.33 | **100.00** | 99.24 | **97.40** |
| Deep(1-4) | 99.38 | 95.70 | 99.00 | 98.55 | 79.50 | 73.20 | 89.87 | 95.00 | **100.00** | 98.86 | 96.78 |
| Stoc | 100 | 96.05 | 99.20 | 98.55 | 83.00 | 73.51 | 91.47 | 95.00 | **100** | 99.81 | --- |
| Hand+Deep | **100.00** | 97.21 | **99.20** | **99.09** | **93.50** | 75.67 | **96.87** | **98.75** | **100.00** | **100.00** | 97.00 |
| Nanni et al. [41], | **100.00** | 95.93 | 98.60 | 98.55 | 91.50 | 75.15 | 90.67 | 94.58 | **100.00** | **100.00** | 93.98 |
| Y. Song et al. [75] | 99.90 | **98.30** | --- | --- | 86.50 | 64.80 | 96.80 | 97.90 | 99.60 | **100.00** | --- |
| Coelho et al. [76] | 98.50 | 94.4 | 95.60 | 88.10 | 67.50 | 44.60 | --- | --- | --- | --- | --- |
| Shamir [77] | 93.00 | 84.00 | --- | --- | 82.00 | 49.00 | 85.00 | 53.00 | 99.00 | 51.00 | --- |
| Zhou et al. [1] | 93.10 | 68.30 | --- | --- | 55.00 | 51.10 | 70.90 | 89.60 | 91.70 | 73.8 | --- |
| Uhlmann et al. [5] | 99.00 | 84.00 | --- | --- | 73.00 | 55.00 | 66.00 | --- | 99.00 | 89.00 | --- |
| B. Zhang & Pham [78] | 98.40 | 90.70 | --- | --- | 90.10 | --- | --- | --- | --- | --- | --- |
| Meng et al. [79] | --- | --- | --- | --- | --- | --- | 92.7 | --- | 99.20 | 96.40 | --- |
| Kather et al. [73] | --- | --- | --- | --- | --- | --- | --- | --- | --- | --- | 87.40 |
| X. Zhang & Zhao [80] | --- | 93.08 | --- | --- | --- | --- | --- | --- | --- | --- | --- |
| Lin et al. [81] | --- | 89.37 | --- | --- | --- | --- | --- | --- | --- | --- | --- |
| Paladini [82] | --- | --- | --- | --- | --- | --- | --- | --- | --- | --- | 96.16 |
| Inés et al. [83] | 98.00 | **99.00** | --- | --- | --- | 73.00 | 95.00 | --- | 100. | 99.00 | --- |

## 5. Conclusion

In this paper, we propose a GenP bioimage ensemble that combines handcrafted and learned descriptors. An ensemble of deep learning methods is built using different criteria (different batch sizes, learning rates, topologies, and methods of data augmentation). We propose three new methods for data augmentation based on feature transforms (principal component analysis and discrete cosine transform) that boost the performance of Convolutional Neural Networks (CNNs). Each handcrafted descriptor is used to train a different Support Vector Machine (SVM), and the different SVMs are combined with the ensemble of CNNs. The experimental section shows that a boost in performance is obtained by combining local features, dense sampling features, and deep learning approaches. The discriminative power and generalizability of our best performing bioimage system are verified on a wide range of publicly available bioimage benchmark datasets, each representing a different classification task.

The main contributions of the proposed paper are the following:

- Three new approaches for data augmentation based on PCA/DCT are presented;
- The power of using different data augmentation techniques for building an ensemble of CNNs is demonstrated;
- A set of handcrafted/learned descriptors is proposed that is not only highly generalizable but also capable of achieving state-of-the-art performance on a large group of diverste image datasets.

In the future, we plan to investigate methods for combining this system with other approaches. We also plan to investigate novel methods for training CNNs on smaller training sets and for reducing the dimensions of deeper CNN layers.

To reproduce the experiments reported in this paper, the MATLAB code for all the descriptors and data augmentation methods is available at https://github.com/LorisNanni.

**Author Contributions:** Author Contributions: L.N. conceived the presented idea., G.M., S.G. and M.P. performed the experiments. S.B., G.M., L.N. wrote the manuscript. S.B. and M.P. provided some resources.

**Funding:** This research received no external funding.

**Acknowledgments:** The authors are grateful to NVIDIA Corporation for supporting this research with the donation of a Titan Xp GPU. We also acknowledge TCSC - Tampere Center for Scientific Computing (Tampere, Finland) for generous computational resources.

**Conflicts of Interest:** The authors declare no conflicts of interest.